\documentclass[runningheads]{llncs}
\usepackage{graphicx}
\usepackage{booktabs}
\begin{document}
\title{Code-Mixed Text to Speech Synthesis under Low-Resource Constraints}
%
%
\author{Raviraj Joshi \and
Nikesh Garera
}
\authorrunning{R. Joshi et al.}
%
\institute{Flipkart, Bengaluru, India \\
\email{\{raviraj.j,nikesh.garera\}@flipkart.com}}
\maketitle              
\begin{abstract}
Text-to-speech (TTS) systems are an important component in voice-based e-commerce applications. These applications include end-to-end voice assistant and customer experience (CX) voice bot. 
Code-mixed TTS is also relevant in these applications since the product names are commonly described in English while the surrounding text is in a regional language. In this work, we describe our approaches for production quality code-mixed Hindi-English TTS systems built for e-commerce applications. We propose a data-oriented approach by utilizing monolingual data sets in individual languages. We leverage a transliteration model to convert the Roman text into a common Devanagari script and then combine both datasets for training. We show that such single script bi-lingual training without any code-mixing works well for pure code-mixed test sets. We further present an exhaustive evaluation of single-speaker adaptation and multi-speaker training with Tacotron2 + Waveglow setup to show that the former approach works better. These approaches are also coupled with transfer learning and decoder-only fine-tuning to improve performance. We compare these approaches with the Google TTS and report a positive CMOS score of 0.02 with the proposed transfer learning approach. We also perform low-resource voice adaptation experiments to show that a new voice can be onboarded with just 3 hrs of data. This highlights the importance of our pre-trained models in resource-constrained settings. This subjective evaluation is performed on a large number of out-of-domain pure code-mixed sentences to demonstrate the high quality of the systems.

\keywords{code-mixed \and text to speech \and encoder-decoder models \and tacotron2 \and waveglow \and transfer learning}
\end{abstract}

\section{Introduction}

Text to Speech (TTS) systems are widely used in voice-based applications \cite{tan2021survey}. These systems are used along with automatic speech recognition (ASR) \cite{joshi2022comparison} to provide an end-to-end voice interface. It is also prominently used in e-commerce applications like  voice assistants, customer experience (CX) voice bots, and user nudges to highlight a feature or product \cite{kraus2019voice,joshi2021attention}. In this work, we describe the approaches used to build the TTS system for e-commerce use cases. 

In a country like India with high linguistic diversity along with English speaking population 'code-mixing' or 'code-switching' is a common phenomenon. With a large Hindi-speaking diaspora, Hindi-English code-mixing is prevalent in social media and e-commerce platforms \cite{nayak2022l3cube}. Moreover with product names and service terminologies mostly described in English an e-commerce voice assistant with Hindi as the primary language should support code-mixing as well. For example, a Hindi sentence \textit{"Mafi chahate hai, par aapke product Babolat Super Tape X Five Protection Tape ko wapis nahi kiya jaa sakta hai"} (We're sorry, but your product Babolat Super Tape X Five Protection Tape is non-returnable) contains product name in English.  We, therefore, focus on building a code-mixed TTS system for e-commerce use cases.   

Building a TTS system requires high-quality studio recordings for training \cite{ning2019review}. It is even difficult to build a code-mixed TTS due to a lack of appropriate data sets, complex methods, and coverage issues. A common approach is to create a mixed-script data set by detecting the language of each word and then transliterating it into the corresponding script \cite{rallabandi2017building,sitaram2016experiments}. The mixed script is preferred as the pronunciation of regional tokens is more accurate in the native script. For Hindi-English code-mixed text, the Hindi words are in Devanagari script whereas the English words are in Roman script. 
Each word is passed to the corresponding language G2P (Grapheme to phoneme) system and the phoneme representations are then passed to the model. An even naive approach is to use a single English G2P model and map Hindi phones to the closest English phones. However, utilizing separate G2P modules for two languages yields good results. These multi-component systems are complex to build and also results in high latency.

\begin{figure}[tb]
  \centering
  \includegraphics[scale=0.4]{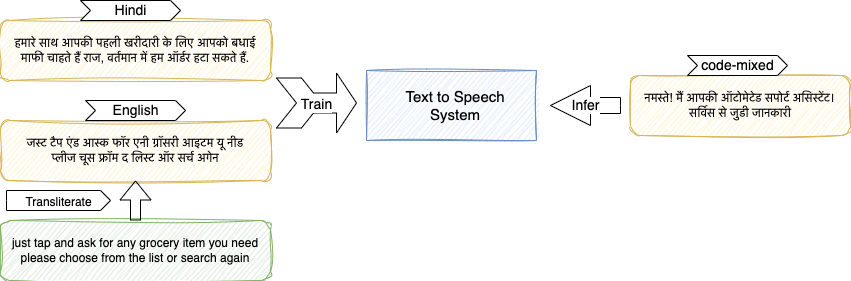}
  \caption{Code-mixed text to speech synthesis}
  \label{fig:comix_tts_flow}
\end{figure}

In this work, we propose a simple data-oriented approach for our use case. Due to a lack of pure code-mixed data, the proposed solution utilizes individual monolingual (text, audio) pairs in Hindi and English. We use an in-house high-quality transliteration system to convert the English data to a common Devanagari script. The Hindi and English data are mixed to train a TTS model converting Devanagari text to speech. Since the primary language of the end application is Hindi we convert all the data to Devanagari script. We show that independent bi-lingual data sets without pure code-mixing work well for pure code-mixed test sets. This approach is shown in Figure \ref{fig:comix_tts_flow}. Although our primary focus is Hindi and Devanagari script with high-quality transliteration systems (English to any Indic Script)  the idea can be easily extended to other languages.

For modeling, we implement a two-stage Tacotron2 + Waveglow architecture \cite{shen2018natural,prenger2019waveglow}. The Tacotron2 model has been used for text-to-spectrogram conversion and the Waveglow then converts the spectrogram into target audio samples. While there are multiple options available for the spectrogram prediction network and audio synthesis network we choose Tacotron2 + Waveglow as they are competitive with other architectures and still popular in literature \cite{favaro2021itacotron,abdelali2022natiq,garcia2022evaluation,finkelstein2022training}. Moreover, there are single-stage end-to-end deep learning models available but these are not considered in this work due to high data requirements. We also present a comparative analysis of single-speaker and multi-speaker Tacotron2 configurations \cite{jia2018transfer}. The single speaker is a standard setup where single-speaker data is used to train the model. In the muti-speaker setup, we utilize speaker embeddings extracted from an external pre-trained speaker verification model to control the output speaker characteristics. A multi-speaker model allows for zero-shot or few-shot voice adaptation and also has the advantage of cross-speaker learning. These approaches are further augmented using pre-training based transfer learning approach. We initially pre-train the single-speaker model by pooling all the speakers together. This pre-trained model is adapted for target speakers in single-speaker configuration and is also used to initialize the multi-speaker model. We show that the single-speaker adaptation of the pre-trained works the best. We observe that although a single multi-speaker model is capable of generating speech for multiple speakers it leads to a slight degradation in the quality of output. These approaches are evaluated using subjective MOS and CMOS scores on a completely out-of-domain test set from the CX domain while the training data is from the Voice Bot and general domain.    
\section{Related Work}

A host of TTS architectures have been proposed over time with a focus on speed and quality. Recently, single-stage fully end-to-end architectures have been proposed which directly convert text to audio samples. These models include VITS \cite{kim2021conditional}, Wave-Tacotron \cite{weiss2021wave}, and JETS \cite{lim2022jets}.  However, these models require a large amount of data. The two-stage models (spectrogram generation + speech synthesis) require comparatively less amount of data as the vocoder can be separately trained with audio-only data. The popular spectrogram prediction networks include Tacotron2 \cite{shen2018natural}, Transformer-TTS \cite{li2019neural}, FastSpeech2 \cite{ren2020fastspeech}, FastPitch \cite{lancucki2021fastpitch}, and Glow-TTS \cite{kim2020glow}. There are a wide variety of vocoders to choose from like Clarinet \cite{ping2018clarinet}, Waveglow \cite{prenger2019waveglow}, MelGAN \cite{kumar2019melgan}, HiFiGAN \cite{kong2020hifi}, StyleMelGAN \cite{mustafa2021stylemelgan}, and ParallelWaveGAN \cite{yamamoto2020parallel}. In terms of voice quality, there is no clear winner among the models and models perform competitively on high-quality datasets. 

These architectures have also been extended to multi-speakers by conditioning them on speaker embeddings. The speaker embeddings encode the speaker characteristics of the target audio. The speaker embeddings are either extracted from an external speaker verification model \cite{jia2018transfer,arik2018neural} or learned jointly during TTS training \cite{pingdeep}. The external embeddings are based on d-vector \cite{wan2018generalized} or x-vector \cite{snyder2018x} systems. The pre-trained and learnable speaker embeddings were compared in \cite{chien2021investigating}. The per-trained embeddings were shown to perform superior performance on FastSpeech 2 model. A similar comparison with the Tacotron model has been performed in \cite{cooper2020zero}. They perform zero-shot speaker adaption using different speaker embeddings but still report a gap between similarity scores of seen and unseen speakers. They observed that these models do not generalize well to unseen speakers. A TTS system incorporating different emotions was studied in \cite{wu2019end}. They use global style tokens (GSTs) to encode the emotion information add the tokens are jointly trained using emotion labels. Similarly, multi-speaker systems using speaker embeddings are also built in \cite{chen2019cross,casanova2022yourtts,cooper2020can}.

Relatively less amount of work has been done in code-mixed TTS systems. A preliminary approach for Hindi-English code-mixed TTS using mixed script text was proposed in \cite{sitaram2016speech}. A language identification system was employed to distinguish Romanized Hindi and English words followed by the transliteration of Hindi words to the Devanagari script. They however used a common English phone set for both Hindi and English words which might result in accent issues for regional words. Further, different Grapheme to Phoneme (G2P) for English words and regional words were utilized in \cite{rallabandi2017building,thomas2018code}. A single mix-lingual G2P model instead of two separate models were proposed in \cite{bansal2020improving}.  In \cite{zhou2020end}, embeddings from an external cross-lingual language model were integrated into the fronted of Tacotron2 model along with the original phone embeddings. The cross-lingual language model encodes words of both languages into the same space thus improving the performance of code-switched TTS. In this work, we make use of a single script and a graphene-based Tacotron frontend thus eliminating the need for such complex high latency modifications. The high quality of the transliteration model also suppresses the pronunciation and accent issues.

\section{Methodology}

\subsection{Code-mixed TTS}
The primary objective of this work is to build a Hindi-English code-mixed or code-switched TTS. Ideally, we would require code-mixed recordings for training such a system. However such recordings are rarely available in practice due to the focus on a single language. To solve for the lack of datasets, we propose a data-oriented approach and utilize monolingual data from the two languages. We use the recordings for Hindi and English text from the same speaker. We propose a single script transliteration-based approach to build a bilingual system. Since the primary language of the end application is Hindi we convert the English text to Devnagari script using an in-house Roman-to-Devanagari transliteration model. The (English text, audio) and (Hindi text, audio) paired data with all the text in the Devanagari script are simply used together to train a single model. We show that this simple mixing works well even for the code-mixed data. We compare Hindi-only training and dual-language training to show the effectiveness of using dual languages. The system is evaluated on 500 strong code-mixed examples from the out-of-domain Customer Experience (CX) domain. We also evaluate the system on a 500 English-only product names test set to showcase the English-speaking capabilities of the model. 

\begin{figure}[tb]
  \centering
  \includegraphics[scale=0.45]{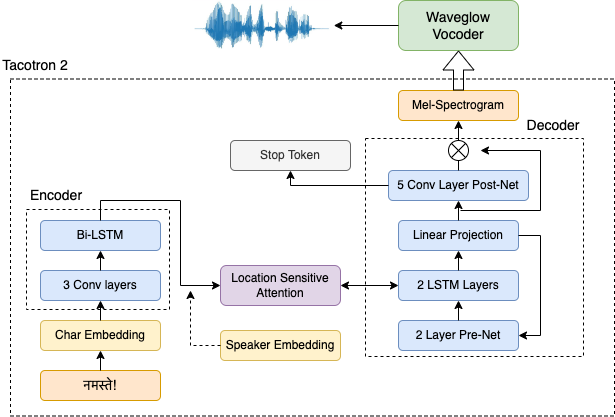}
  \caption{Model Architecture for single-speaker and multi-speaker configurations. The multi-speaker model has an extra speaker embedding component extracted from a pre-trained speaker verification model.}
  \label{fig:taco_model}
\end{figure}

\subsection{Model Architecture}
In this work, we use the Tacotron2 spectrogram prediction network and Waveglow vocoder for all our data-oriented experiments. We perform both single-speaker and multi-speaker experiments. For the multi-speaker model, we simply fuse the external x-vector speaker embeddings with the Tacotron model. The model architecture is described in Figure \ref{fig:taco_model}. In the next sub-sections, we describe network architecture and the experimental setup.
\subsubsection{\textbf{Single-Speaker Tacotron2}}

The Tacotron2 \cite{shen2018natural} is an auto-regressive encoder-decoder model that maps text sequence to spectrogram sequence. We use characters as input to the encoder and the architecture is the same as that described in the original work. The encoder consists of three Conv layers followed by a Bi-LSTM layer. The character embedding size, number of Conv filters, and Bi-LSTM units is 512. The Conv filter size is 5 x 1. The decoder is an auto-regressive network conditioned on encoder output. It uses location-sensitive attention \cite{chorowski2015attention} to compute the context vector. It consists of two uni-LSTM layers with 1024 units. The decoder also consists of a pre-net and post-net added before and after the uni-LSTM layers respectively. The pre-net consists of 2 feedforward layers (256 units) and the post-net consists of 5 Conv layers (512 filters with a size of 5x1). The output of the uni-LSTM is concatenated with the context vector and is passed through two parallel dense layers to compute the stop token and target log-mel spectrogram. The spectrogram is further refined using the post-net and a residual connection connects the output spectrogram to the output of the post-net. The mean squared error (MSE) loss is used for training.

\subsubsection{\textbf{Multi-Speaker Tacotron2}}
This model is the same as the single-speaker model except for the addition of speaker embeddings. An x-vector system is used to extract the speaker embeddings \cite{snyder2018spoken} from the corresponding audio sample. The pre-trained model \footnote{https://huggingface.co/speechbrain/spkrec-xvect-voxceleb} is based on a time-delay neural network (TDNN). It was trained using the VoxCeleb speaker recognition dataset. The 512-dimensional embeddings are subjected to LayerNorm followed by a dense layer of size 512. The output of the dense layer is again passed through a LayerNorm and added to each time step of the encoder output. The decoder is therefore conditioned on the speaker embeddings as well in order to generate audio for the desired speaker. The multi-speaker model again has two configurations as described below.
\begin{itemize}
    \item \textbf{Audio embedding}: This is the regular configuration in which speaker embedding for each audio is computed at run time and passed to the model. The audios from a specific speaker are not explicitly distinguished. We observe that this model shows some generalization to un-seen speakers however at times fails to generate an end token on some audio samples during inference.
    \item \textbf{Avg embedding}: This configuration is similar to speaker selection where each speaker is assigned a single speaker embedding which is the average of speaker embedding from all the audios of the corresponding speaker. With this configuration, we do not see the end token issue however this does not work for unseen speakers.
\end{itemize}

\subsubsection{\textbf{Pre-training strategies}}
We explore transfer learning from public LJSpeech English data and all the available Hindi data from multiple speakers. We observe that pre-training strategies are essential for building a high-quality model. The following strategies are followed for both single-speaker and multi-speaker models.
\begin{itemize}
    \item \textbf{English warmstart}: The Tacotron2 model is initially trained on English LJSpeech corpus with the input in Roman script. During target speaker fine-tuning the character embedding layer has to be discarded since our experiments are based on the Devanagari script.
    \item \textbf{Mix-data warmstart}: In this setup, we initialize the model with weights from English training and then further train the entire model on a mixture of all the speaker's data in the Devanagari script. This model is not directly useful since it is a single-speaker model trained with multi-speakers. This model will generate different speakers' voices for different sentences and is typically biased toward one specific speaker. However, a rich text encoder obtained from this mixed training hence acts as a very good initialization for target speaker adaptation.
\end{itemize}

\subsubsection{\textbf{Fine-tuning strategies}}
Based on the pre-training strategies we follow two fine-tuning methods.
\begin{itemize}
    \item \textbf{Full fine-tuning}: The english-warmstart models are subjected to full-finetuning. This is required because of the mismatch in the script for the English text and Devanagari text.
    \item \textbf{Decoder only fine-tuning}: The mix-warmstart models the encoder is already trained on a large amount of Devanagari text. So we freeze the encoder parameters and perform decoder-only fine-tuning. While we can perform full finetuning with mix-warmstart model it gives slightly lower performance than decoder-only finetuning.
\end{itemize}
We perform an ablation study with pre-training and fine-tuning methods to show that mix-warmstart + decoder-only finetuning works the best.

\subsubsection{\textbf{Low-resource voice adaptation}}
We perform low-resource voice adaptation experiments in order to understand data requirements for onboarding a new voice/speaker. We use the pre-trained models and perform single-speaker adaptation using different low data configurations like 3 hrs, 5 hrs, and 10 hrs. We observe that 3 hrs of data is sufficient to get a high-quality model with mix-warmstart models. The experiments corresponding to 3 hrs of data are reported in this work. Recently, a TTS system Vall-E \cite{wang2023neural} has shown extraordinary zero-shot capabilities. However, this system uses a complex architecture and requires 60K hours of pre-training data making it infeasible in low-resource scenarios. Our work uses data of order 15 hours and therefore cannot be compared with such system utilizing 60k hours of data.

\subsubsection{\textbf{Waveglow Vocoder}}
The waveglow \cite{prenger2019waveglow} model converts mel-spectrogram into audio samples. It is a flow-based generative model which generates audio samples by sampling from a distribution. It performs a series of invertible transforms to convert examples sampled from zero mean and spherical Gaussian distribution into target audio samples. The transformation is also conditioned on mel-spectrogram. The model minimizes the log-likelihood of the data. 

The mel-spectrogram is computed using short time fourier transform (STFT). It uses a frame length of 50 ms and a hop size of 12 ms. An 80-channel Mel filter bank is used to transform STFT into Mel scale followed by log compression.

\subsection{Dataset Details}
The datasets used in this work are described below.
\begin{itemize}
    \item \textbf{English LJSpeech Corpus}: It is a publicly available single-speaker corpus consisting of 13100 (text, audio) pairs \cite{ljspeech17}. The text is taken from 7 non-fiction English books and the total size of the data is 24 hours.
    \item  \textbf{Single-Speaker Data}: This is an in-house studio recording from a female speaker. The text for the recordings is taken from the Voice assistant domain and general domain (Wikipedia-like sentences). The total size of the data is 15 hrs consisting of both English and Hindi text. Roughly 65\% of the data is Hindi and the rest is English. This speaker is also used for the evaluation of all the models explored in this work. 
    In order to perform low-resource voice adaptation experiments we choose a random subset of 3 hours from this data-set.
    \item \textbf{Multi-Speaker Data}: We further create a multi-speaker corpus using additional 4 speakers including 2 male and 2 female speakers. The text from the above single-speaker data is used for recording. The size of data for each speaker is approximately 15 hours. These 4 speakers along with the above primary speaker are used for multi-speaker training. The primary female speaker is also chosen for testing the multi-speaker models. Since we use only 5 speakers in multi-speaker training the generated samples are highly similar to the original speaker. Hence similarity tests are not reported in this work. The goal of this work is to create a high-quality primary speaker system and all the evaluations are designed with this objective.
    \item \textbf{Test datasets}: We create two out-of-domain test sets for the evaluation of all the models. These test sets are from customer experience (CX) and product domains. The CX test set consists of customer queries and bot responses with high Hindi-English code-mixing.  The product test set consists of English product names from e-Commerce listings. The size of both test sets is 500 text examples. The audios were synthesized for these texts from the corresponding model and used for MOS (Mean Opinion Scores) and CMOS (Comparative MOS)  evaluation. For CMOS evaluation, we use the output from out-of-the-box Google speaker 'hi-IN-Standard-A' as the reference audio.  
\end{itemize}

\begin{table*}
  \caption{Subjective MOS scores for different model configurations on code-mixed CX test set}
  \label{tab:mos_res}
  \begin{tabular}{p{6.5cm}p{2.5cm}p{3cm}}
    \toprule
    TTS Type & eng-warmstart (full train) & mix-warmstart (decoder only train)  \\
    \midrule
    single speaker (Hindi only) & 4.37  +- 0.78 & 4.36  +- 0.86 \\ \hline
    single speaker (Hindi + English) & 4.58  +- 0.69 & \textbf{4.65 +- 0.56} \\ \hline
    multi-speaker (audio-embed, Hindi + English) & 4.55  +- 0.64 & 4.65  +- 0.50 \\ \hline
    multi-speaker (avg-embed, Hindi + English) & 4.44  +- 0.95 & 4.61  +- 0.7 \\ \hline
  \bottomrule
\end{tabular}
\end{table*}

\begin{table*}
  \caption{Subjective CMOS scores for different model configurations. All the rows except for the last row corresponds to CX test set. The last row indicates numbers for Product test set.}
  \label{tab:cmos_res}
  \begin{tabular}{p{6.5cm}p{2.5cm}p{3cm}}
    \toprule
    TTS Type & eng-warmstart (full train) & mix-warmstart (decoder only train)  \\
    \midrule
    single speaker (Hindi only) & -0.47 & -0.35 \\ \hline
    single speaker (Hindi + English) & -0.45 & \textbf{0.02} \\ \hline
    multi-speaker (audio-embed, Hindi + English) & -0.42 & -0.3 \\ \hline
    multi-speaker (avg-embed, Hindi + English) & -0.8 & -0.14 \\ \hline \hline
    single speaker (Hindi + English, Product test set) & - & \textbf{0.12} \\ \hline
  \bottomrule
\end{tabular}
\end{table*}

\begin{table}
  \centering
  \caption{MOS scores for low-resource voice adaptation experiments on target speaker}
  \begin{tabular}{cc}
    \hline
    \textbf{Training Strategy} & \textbf{MOS}\\
    \hline
    eng-warmstart + Target (3 hrs) & 4.27 $\pm$ 0.95 \\ \hline
    mix-warmstart + Target (3 hrs) & 4.54 $\pm$ 0.58\\ \hline
    mix-warmstart + Target (frozen encoder, 3 hrs) & \textbf{4.59} $\pm$ 0.68  \\ \hline
    mix-warmstart + Target (frozen encoder, 15 hrs)& 4.65 $\pm$ 0.58 \\ \hline
\end{tabular}
  \label{tab:low_res_adapt}
\end{table}

\begin{figure}[htb]
  \centering
  \includegraphics[scale=0.25]{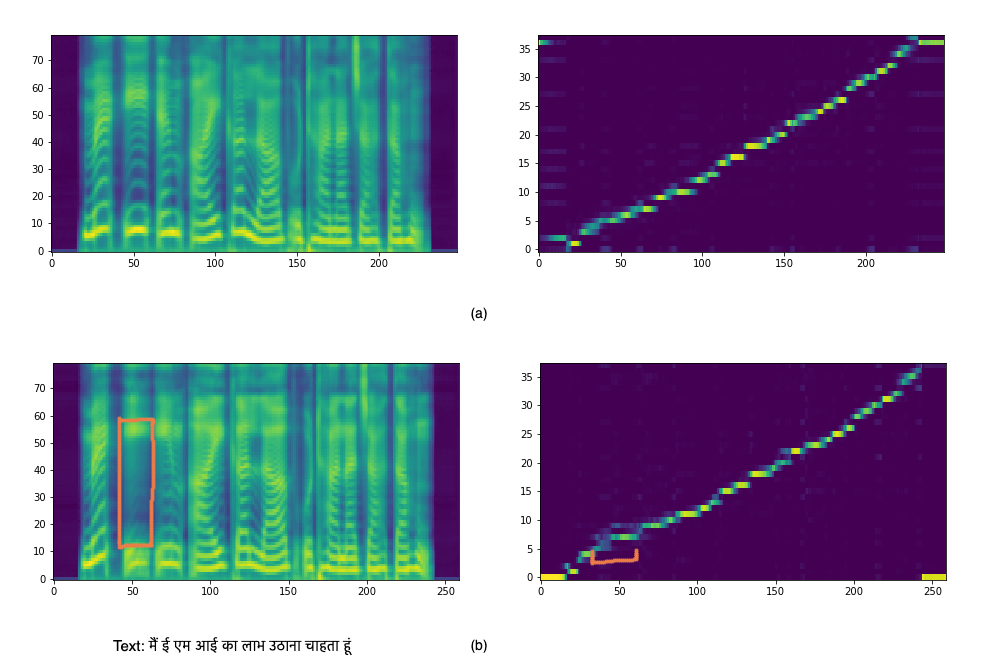}
  \caption{The Mel-spectrogram and attention alignment plot for a sample sentence using config (a) Mixed [Hindi + English] Training (b) Hindi only training. The difference in the resolution can be clearly seen at the start of the two spectrograms at the word EMI. }
  \label{fig:tts_spectrogram}
\end{figure}

\section{Results}
In this work, we evaluate different single-speaker and multi-speaker TTS models for code-mixed speech synthesis tasks. Independent Mean Opinion Scores (MOS) and comparative CMOS scores are used to compare these models. These evaluations are done by 50 trained individuals with each listener evaluating around 30 audios. The audios are presented in random order and specifically, during CMOS the reference audio is randomly chosen. In MOS evaluation the listener is asked to rate the audio on a 1-5 (with a gap of 0.5) scale. A score of 5 indicates a naturally sounding voice with perfect pronunciation. The naturalness and pronunciation are evaluated during MOS, the higher the score better the system. 

In the CMOS evaluation user listens to both audios from our TTS and Google TTS. They provide a rating to the second audio in the (-2 to +2) range. Again based on the naturalness and pronunciation of the second audio is given a +ve rating if it is better than the first audio. A score of 0 indicates that both systems are equally better. The speaker for the first and second audios are randomly selected and scores are internally adjusted such that a +ve rating indicates our speaker is better as compared to the Google speaker and a -ve rating indicates vice versa. The reported MOS and CMOS scores are average of all the individual scores. Standard practices are followed to avoid listener fatigue and bias.

All the models are evaluated in two configurations eng-warmstart and mix-warmstart. The eng-warmstart indicates English data pre-training with full fine-tuning whereas mix-warmstart indicates mix data pre-training with decoder-only finetuning. The MOS results are shown in Table \ref{tab:mos_res} and the CMOS scores are shown in Table \ref{tab:cmos_res}. We show that (Hindi + English) training works better than Hindi-only training via both MOS and CMOS scores. A sample spectrogram for the two configurations is shown in Figure \ref{fig:tts_spectrogram}. 

The mix-warmstart configuration shows clear improvements over eng-warmstart thus highlighting the importance of Devanagari-based pre-training. While comparing multi-speaker and single-speaker models, the single-speaker based adaptation works better. The difference is more prominent in comparative CMOS score as compared to the MOS score. A positive CMOS score for the single-speaker mix-pretrained model indicates that the system is slightly better than the Google system on code-mixed test sets. Finally, while comparing the two speaker embedding methods for multi-speaker models there is no clear winner. The CMOS scores are in favor of avg-embed whereas MOS scores are in favor of audio-embed. We personally felt that audio-embed systems are slightly better. We also perform a CMOS evaluation of the best single-speaker system on the English product names test set. A high +ve CMOS score indicates that the dual data training is also helping the model beat the Google system.

The results of low-resource speaker adaptation are described in Table \ref{tab:low_res_adapt}. We observe that the mix-warmstart models can be adapted to a new speaker using just 3 hours of data. We again use decoder-only fine-tuning in this setup. The degradation in MOS scores is very less even after using just 1/5th of the original data. 

\section{Conclusion}
We present different approaches utilized to build a production-quality TTS system for code-mixed e-commerce use cases. We propose a transliteration-based approach to convert the dual language data into a common script and use it for training. We show that this dual language training also works well for code-mixed test sets. We compare different single-speaker and multi-speaker TTS models using two different pre-training methods. We show the advantages of transfer learning from the mix-pretraining setup. The multi-speaker models are further evaluated in reference audio (audio-embed) and speaker selection (avg-embed) configurations. The single-speaker model with mix-data pre-training performs the best and it is also shown to perform better than the Google TTS on code-mixed use cases. We also show that the mix-data pre-trained models with decoder-only fine tuning can be adapted to a new voice with just 3 hours of data. This shows the importance of pre-trained models in a low-resource setting.

\bibliographystyle{splncs04}
\bibliography{main}
%




\end{document}